\documentclass{article}

\usepackage{PRIMEarxiv}
\usepackage[utf8]{inputenc} 
\usepackage[T1]{fontenc}    
\usepackage{hyperref}       
\usepackage{url}            
\usepackage{booktabs}       
\usepackage{amsfonts}       
\usepackage{nicefrac}       
\usepackage{microtype}      
\usepackage{lipsum}
\usepackage{fancyhdr}       
\usepackage{graphicx}       

\usepackage{amsthm}
\usepackage{amsmath}
\allowdisplaybreaks[4]
\usepackage{mathrsfs}
\usepackage{amssymb}
\def\d{\mathrm{d}}
\def\I{\mathrm{I}}
\def\v{\mathbf{v}}
\def\P{\mathbb{P}}
\def\R{\mathbb{R}}
\def\E{\mathbb{E}}
\def\scrX{\mathscr{X}}
\def\scrY{\mathscr{Y}}

\def\R{\mathbb{R}}
\def\B{\mathbb{B}}
\usepackage{relsize}
\usepackage{braket}
\usepackage{enumitem}

\newtheorem{theorem}{Theorem}[section]
\newtheorem{lemma}[theorem]{Lemma}

\newtheorem{prof}[theorem]{Proof}
\newtheorem{proposition}[theorem]{Proposition}

\newtheorem{remark}[theorem]{Remark} 
\newtheorem{example}[theorem]{Example}
\newtheorem{hypothesis}[theorem]{Hypothesis}

\title{Probabilistic Results on the Architecture of Mathematical Reasoning Aligned by Cognitive Alternation
}

\author{
  Minzheng Li, Xiangzhong Fang, Haixin Yang \\
  School of Mathematical Sciences, Peking University \\
  \texttt{1801110057@pku.edu.cn} 
 }

\begin{document}
\maketitle

\begin{abstract}
We envision a machine capable of solving mathematical problems. Dividing the quantitative reasoning system into two parts: thought processes and cognitive processes, we provide probabilistic descriptions of the architecture.
\end{abstract}


\section{Introduction}
AlphaGo has made ground-breaking establishment in the large searching space problems at the game of Go\cite{Silver2016MasteringTG}\cite{silver2017mastering}. This is followed by ChatGPT earlier this year, gaining attraction and popularity among both individuals and scientists \cite{gao2023comparingGPT1}\cite{sorin2023largeGPT2}\cite{emmert2023canGPT3}\cite{van2023chatgptGPT4}\cite{takefuji2023briefGPT5}\cite{duong2023analysisGPT6}. We are also gratified to witness the involvement of artificial intelligence in scientific research assisting humans, summarized in the latest Nature article\cite{wang2023scientific}. It is time for us to undertake the task of building a machine that is capable of solving mathematical problems and exercises. 

Google Research has posted two versions of pre-prints on machines of mathematical reasoning\cite{lewkowycz2022solving}.

\cite{OpenAIwebMathReasoning} records the recent work by OpenAI on mathematical reasoning. Large language models have made significant progressing in multi-step reasoning, but they still produce logical mistakes. Researchers in OpenAI apply supervision process to reduce mistakes.

The Baidu company with their developing yiyan \cite{BaiduwebMathReasoningyiyan} as well aims for mathematical reasoning. 

The OpenAI's blog is one of the most important inspirations of our article.

The thought process in mathematics is expressed through mathematical language. In order for a mathematical machine to function effectively, it must be able to identify and utilize the correct mathematical language. For machines solving mathematical problems, ensuring the accuracy of mathematical language is primary. The inspector coexists and cooperates with the generator of mathematical language. The mathematical machine needs to inspect the language it generates, as well as possibly generating intermediate processes to inspect a mathematical statement correct or not. We highly regard the current popular large language models as the most plausible option for inspecting the accuracy of mathematical content, and we mainly in this text discuss the comprehension mechanisms involved in generating mathematical text.

Mathematics is expressed as a language. We offer several observations of vocabulary by reflecting on linguistic models and systems. We have magnificent terms like "centennial" in English, matching with comprehensive concepts such as "linear vector space" in mathematics. Each language also has comparatively trivial expressions, such as "of a hundred years" in English, or the eight axioms embodying the definition of linear spaces in mathematics. Languages vary in complexity, ranging from intricate and succinct to simple and straightforward. We specifically want to mention the C programming language as a noteworthy phenomenon. The C programming language is considered as the efficient "big words", while the binary execution codes serve as the "trivial words". The C compiler functions as a translator, interpreting the human-like language into the format executable by machines. The technology is a great invention, pity for us to often take its existence for granted. Being capable of interchangeably using all levels and formats of mathematical language is considered a fundamental aspect of understanding mathematics.

Cognitive psychology textbooks, such as those by \cite{anderson2005cognitive},\cite{galotti2017cognitive},\cite{goldstein2014cognitive}, offer fundamental knowledge in the field of cognitive science. An individual's memory capacity largely depends on their familiarity with the system's structure. For instance, professional chess players can recall sixteen positions with a single five-second glance at the board, while amateur players can remember only five or six. However, when the chess pieces are randomly placed, both professional and amateur players can recall only two or three positions by a glance. To improve efficacy of a mathematical machine in its thought processes in the working memory, cognitive training is expected to not only fine-tune parameters but also facilitate the machine's self-construction in understanding mathematical systems and structures. The machine's comprehension is to be able to become familiar with a wide range of mathematical structures and phenomena, thus to talk and reason with certain background or from certain concepts.

Let's examine several mathematical problems and offer my subjective advice for each, where we review some core comprehensive characteristics in the thought processes of solving mathematical exercises and problems:
\begin{example}[Solution by pure deduction: Stein's lemma]
Prove Stein's lemma $$
\E(\triangledown g(X)) =\Sigma^{-1} \E(g(X)(X-\mu))
$$
for $ X \sim \text{normal}(\mu,\Sigma) $. Solution: the exercise is solved by a simple one-step deduction of integration by parts. The Gauss-Green theorem which is multi-dimensional generalisation of integration by parts, and the idea differentiation operator being symmetric operator in real inner products, are possibly prompted up in the mind. \qedsymbol
\end{example}
\begin{example}[Solution by proof of contradiction] For every subsequence $ \{X_{n_k}\} $ of $ \{X_n\} $, there exists further subsequence $ \{X_{n_{k_l}}\} $ that converges to $ X $ almost surely, prove that $$
X_n\to X\text{ in probability}
$$
This is equivalent condition of convergence in probability. You need to persist into the credit of proof by contradiction: not converging in probability includes not converging in $ L_1 $, there exists $ \epsilon_0 $ such that $$
\E\|X_n - X\| \geqslant \epsilon_0
$$
However $$ \E\left(\min\left\{\|X_n - X\|,1\right\}\right) \geqslant \epsilon_0 $$ contradicts with the dominated convergence theorem since $ \lim\limits_{k\to\infty} \E\left(\min\left\{\|X_{n_k} - X\|,1\right\}\right) = 0 $.\qedsymbol
\end{example}
\begin{example}[Solution by conception: Existence and uniqueness of solutions of ordinary differential equations] To prove $$
\frac{\d}{\d t}x(t) = \v(t,x(t))
$$
having unique solution, given $ \v $ being Lipschitz continuous of its second variable, we construct the idea of Picard iteration sequence $$
\phi_{j+1}(t) = x(0) + \int_0^t \v(s,\phi_j(s))\d s
$$
and show that $ \{\phi_j\}_{j=1}^\infty $ is Cauchy sequence to prove the theorem.\qedsymbol
\end{example}
\begin{example}[Solution by supplementing additional items]
The Chebyshev-type inequalities, including the Markov inequality for non-negative random variables, is proved by supplementing an additional item that is greater than $ 1 $ into the expectation.\qedsymbol
\end{example}
\begin{example}[Solution by the convergence or divergence of series]
Very many mathematical problems come down to the convergence or divergence of real number series. For example, the contraction mapping theorem is proved because the number series $ \sum_{j=1}^\infty \alpha^j $ converges when $ \alpha \in (0,1) $.\qedsymbol
\end{example}
\begin{example}[Solution by inequality: the convergence of Q-learning]
In reinforcement learning, $ \gamma $ is the discount factor. $ s $ is the state, $ a $ refers to the action, $ s^{'} $ is the random next state after $ s $. $ r $ is the reward function. Q-learning$$
Q(s,a) \longleftarrow Q(s,a) + \alpha\left( r(s,a) + \gamma \sum_{s^{'}} p(s,a;s^{'})\max_{a^{'}} Q(s^{'},a^{'}) -Q(s,a)\right)\\
$$
converges the the optimal Q function $ Q^* $:$$
Q^*(s,a) = r(s,a) + \gamma \sum_{s^{'}} p(s,a;s^{'})\max_{s^{'}} Q^*(s^{'},a^{'})
$$
The solution is obtained by constructing an inequality to prove $$
\mathcal{L} Q\triangleq (1-\alpha)Q(s,a) + \alpha\left( r(s,a) + \gamma \sum_{s^{'}} p(s,a;s^{'})\max_{a^{'}} Q(s^{'},a^{'}) \right)
$$
is contraction mapping.\qedsymbol
\end{example}

Human psychological processes and linguistic reasoning exist in very difference, although linguistic reasoning is part of human psychology. It is not uncommon to witness instances where individuals disrespect others or disregard facts about the world\cite{peck2002road}. Through a series of smart research experiments\cite{40cite20230712_1}, modern psychology continuously reminds us of our inherent irrationality, where the psychology comes around, contrasting with our linguistic rationality. Ultimately, our linguistic reasoning is an inherent aspect of our psychological nature\cite{wiener2019cybernetics}[Chapter 5].
\begin{example}[Method of mathematical induction]
\leavevmode 
\begin{itemize}
\item $ P(n) $ is a proposition of positive integer $ n $.
\item $ P(1) $ is true.
\item $ P(n) $ is true includes $ P(n+1) $ being true.
\item $ P(n) $ is true for all integers $ n \geqslant 1$.
\end{itemize}
\end{example}
\noindent Let us explore the concept of mathematical induction. The reason why mathematical induction holds lies in the fact that we initially accept this method and then proceed to develop logical frameworks to prove its validity. The logical frameworks that prove mathematical induction is, together with mathematical induction, part of our psychology. 

Our brains encompass more than comprehension and reasoning, but also sensitive thinking. In fact, \cite{anderson2005cognitive}[Chapter 10] introduces two psychological experiments revealing that people are sensitive thinkers but rather suck reasoners, and with the more sensitive information given, the less people reason. On the other hand, sensitive thinking such as intuition do help us in doing mathematics. You must have doubted that a circle encloses the largest area in your primary school, far before finally proving it in university with calculus of variation or other techniques. AlphaGo was designed by mimicking human thinking by Monte Carlo tree search\cite{Silver2016MasteringTG} and by its self-established policy scheduler\cite{silver2017mastering}, but without mimicking mechanisms of sensitive thinking. We have witnessed the remarkable success of Go AI, as they have evolved to become coaches for humans, offering assistance with thousands of josekis (open and established patterns by both players which is considered fair) and ideas. In this text, we avoid discussing the possibility of mechanisms of sensitive thinking.

Mathematical reasoning can be seen as an uninformed, self-constructed search for logical deductions guided by a comprehensive understanding of mathematical concepts and methods. We need to mention that the mouse-maze system as typical and fundamental in scientific and technological methods of searching \cite{shannon1993presentation}\cite{caruso2016fast}. Maze-solving competitions, which remain popular among young people worldwide, often modify the fundamental algorithms of Depth First Search and Breadth First Search. In competitions, BFS is less efficient by setting the mouse move back and forth. AlphaGo \cite{Silver2016MasteringTG}, a groundbreaking Go program, achieves reasoning in the game of Go through Monte Carlo Tree Search. In this text, we mimic Depth First Search and Breadth First Search to discuss possible genres of 
reasoning that delve into one thread of thoughts as deep as possible, as well as multi-threaded workings to explore potential interactions among different ideas. To understand what it means for s sparse search space, try to make a word from letters "TTICAMHEMAS". Brute force search covers titanic number of possibilities, while reflecting that this might have relation to certain scientific concepts in an scientific article as you are reading would help.

Reasoning can be seen as the internal control process in the brain, while control refers to the external, machine-coded reasoning in its operational procedures. It is worth noting that Weiner scratched into the field of artificial intelligence from theories of control in his masterpiece\cite{wiener2019cybernetics}. Nowadays, artificial intelligence and control theories have evolved into distinct disciplines, with mathematical theories addressing typical problems of different backgrounds. A difference is there exist theories of optimal control, employing methods like calculus of variations and the Hamilton-Jacobi-Bellman equation to analyze the spatio-temporal distribution of a system's value, it is uninteresting to define or discuss optimal reasoning. Mathematical students are required to get more than sixty scores to pass the text, without emphasis on the reasoning strategies employed. "Students remember you need to establish yourself by your own," said an old teacher who never checks students' attendance in class. On the other hand though, we appreciate elegant solutions to mathematical problems, where those that leverage classical structures or exploit sufficiently characteristics of the problems are particularly enjoyable. Describing motion by ordinary differential equations$$
\frac{\d }{\d t}x(t) = \v(x(t),\theta(t)) 
$$
where $ \theta(t) $ is the controlling parameter process and $ x(t) $ is the motion process, the optimal control is usually denoted as $ \theta^*(t) $. In this text, we describe single-threaded reasoning, borrowed from search theories by the name depth-oriented reasoning, by ordinary differential equation $$
\frac{\d }{\d t}x(t) = \v(x(t),y) 
$$
where $ y $ is the underlying cognitive state to align with.

The authors of this article have noticed the success of the SOAR cognitive architecture in the field of cognitive science. The SOAR architecture divides its internal structure into working memory and generative memory, with the former mimicking human's consciousness and the latter mimicking human's long-term memory and knowledge. To make a clever reference from our discussion to the SOAR, the thought processes as consciousness are situated in working memory, while cognitive processes related to comprehension primarily operate within generative memory. Inspired by the SOAR cognitive architecture, the working memory is aligned by but unconscious of the operating of cognitive backstage which resides in the generative memory.

The architecture mimics human psychology by incorporating both cognitive and deductive reasoning systems. The cognitive system, which is self-established, consists of a linguistic framework that provides descriptive mathematical results (such as the existence and uniqueness of solutions to ordinary differential equations), definitions (such as linear spaces and operators), and conceptions (such as the technique of Picard iteration sequence). It should also include heuristics for possible direction of evolution of thought processes (for example, applying the Chebyshev type inequality). A true understanding of mathematics should also include self-establishment, the cognitive system refines itself through learning and practicing mathematics. 

The thoughts $ x(t;\omega)\in\scrX $ evolve in alignment with the cognition $ y(t;\omega)\in\scrY $. The cognition $ y(t;\omega) $ renew itself with stochasticity within stochastic time length, which is set to be exponential distribution in our text. We study the statistical behavior of this phenomenon, by deducing the partial differential equation and integral equation of the probability density. This is the content of sections of Depth-Oriented Reasoning and Proof of the Theorem. In our discussion of single-threaded reasoning, the system of reasoning is divided by the thought processes $ x(t) $ in the thought space $ \scrX $ and the cognitive processes $ y(t) $ in the cognitive space $ \scrY $, while in the multiple-threaded of a batch of $ x(t) $'s and $ y(t) $'s at the end of this article, the division is the thoughts state $ \psi(t) $ and the cognitive state $ \phi(t) $.

In classical probability theory, the probability density describes the likelihood of presence of a random variable with respect to the Lebesgue measure. In quantum mechanics, the density matrix $ \rho $ describes a projection from an arbitrary quantum vector $ \ket{u} $ to the quantum system of $ \rho $. Although labeled with density, they are actors in two theaters.

As mathematicians we are compelled to work in this field, due to the fact that our pens and papers are much cheaper than conducting one computer experiment by large companies. Mathematicians engage with the computer community and enhance computer scientists' problem-solving abilities by conceptualizing potential algorithms and architectures, verifying the mathematical forms, quantifying the complexity and the capacity of an architecture, and etc.. Every work is better done by mathematicians than computer experiments. Yet, our motivation goes beyond budget consideration. In the culture of our country, there's a saying: "Unfortunate for the nation, fortunate for the poets." Challenging times in the field of machine learning and artificial intelligence present opportunities for mathematicians to build achievements. We dream of having the possibility of enveloping and sealing the whole ChatGPT into a laptop, with which singular value decomposition for the linear operators seems to not go afar. We admire mathematical works that bridge between theory and practice to lead to clearer thinking. The three authors of this article are two Ph.D. students in statistics and a professor in statistics. In statistical works, we exploit the nature of problems to bring presumptions, as well as deducting assumptions to make the theories more applicable; though these results in the conflict of more or less premises in the forms of mathematics. Assuming the thought processes move in an velocity field is certainly a bug, yet this is considerably basic for further development of mathematical and statistical theories, longing for mathematical modeling that is more adaptive, neat, and encompassing.

\section{Depth-Oriented Reasoning}
In statistics we long for the optimal architecture that interconnect information, memory, and knowledge in mathematical reasoning. In our article, we mimic the idea from OpenAI's blog, and give a discussion of the equations of the probability density.

\begin{remark}[Notation Convention]
$ \scrX $ is the thought space where ideas are written out to be further developed. $ \scrY $ is the established mathematical knowledge and comprehension, which the machine aims to further construct by itself. $ x(t) $ and $ x(t;\omega) $ denote stochastic processes. We abuse the symbol and use $ x $ to denote am element in the space of $ \scrX $. It should be noted that $ x $ is irrelevant of $ x(t) $, but the usage is an abuse of symbol. $ y(t) $ and $ y(t;\omega) $ denote stochastic processes. We abuse the symbol and use $ y $ to denote an element in the space of $ \scrY $. $ x(t) $ and $ y(t) $ denotes motions in the thought space (probably working memory) $ \scrX $ and the cognitive space (probably generative memory) $ \scrY $. $ x $ and $ y $ are positions in $ \scrX $ and $ \scrY $. $ x(t;\omega) $ is continuous motion, and $ y(t;\omega) $ is stochastic jump processes. We dumbly use the symbol $ \tau $ to denote both the random variable and its integrand variable. $ \lambda $ is the Poisson renewal rate. For any $ \omega\in\Omega $, $ y(\cdot;\omega) $ took a jump at time $ t-\tau(t,\omega) $ and did not took jump in $[t-\tau(t,\omega),t]$. $ \tau $ is the last time before $ t $ that $ y(\cdot) $ took its alternation. $ B(x,\epsilon) $ is the $ \epsilon$-radius ball centred at $ x $, and $ |B(x,\epsilon)| $ is to take Lebesgue measure. $ B(x,\epsilon) $ and $ B $ are two symbols.
\end{remark}

\begin{hypothesis}
The evolution of thought processes in $ \scrX $ is depicted as motion within a velocity field in this text. The autonomous motion of $ x(t) $ is induced by the cognitive state $ y $: $$
x(t+1) = x(t) + \v(x(t);y)
$$
illustrating one-step evolution of thought. In our study of mathematical reasoning, for most of the time we observe a lack of guidance to thought processes by its aligning cognitive states, leading to no evolution of thought: $$
\v(x;y) = 0\text{ for most }x\in\scrX\text{ and most }y\in\scrY
$$
Therefore, renewing cognition, referred to as cognitive processes or cognitive states is necessary for the continuation of reasoning. The architecture is characterized in the following paradigm:$$
\mathlarger{\mathlarger{\scrY}}\mathop{ \mathop{}_{\overset{\v}{\longrightarrow}}^{\overset{\psi}{\longleftarrow}} }\mathlarger{\mathlarger{ \scrX}} 
$$
$ y(t) $ supervise and induce $ x(t) $ via the velocity field $$
\v(x,y)\in C^1(\scrX\times\scrY;\scrX)
$$
and we need to keep renewing $ y(t) $. $ x(t) $ is a continuous motion characterized by ODE, and $ y(t) $ is the stochastic jump processes renewed.
\end{hypothesis}

\begin{hypothesis}
We assume the theorem of existence and uniqueness of ordinary differential equations holds for $ x(t;\omega) $ in the velocity fields.
\end{hypothesis}

\begin{figure}[htb]
\centering
\includegraphics[width = 0.4\textwidth]{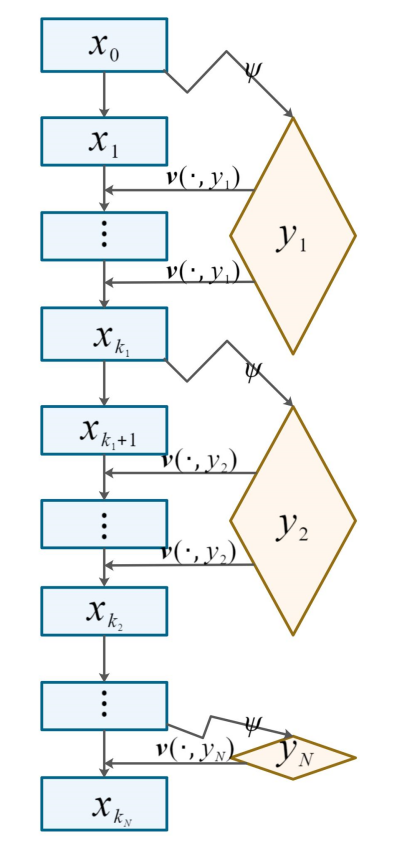}
\caption{Illustration of the single-threaded reasoning}
\label{Fig20230817_1}
\end{figure}

The probability space of stochastic processes $ x(t;\omega),y(t;\omega) $ is $$
\Omega\times\R\to\scrX\times\scrY\qquad\mathrm{or}\qquad\Omega\times\R^+\to\scrX\times\scrY\
$$
depending on whether time starts at $ 0 $. The single-threaded thinking process\begin{equation}\label{Depth_O_thinking20230723}
\begin{split}
&[x_0,y_0]\rightarrow[x_1,y_1]\rightarrow[x_2,y_1]\rightarrow\cdots\rightarrow[x_{k_1},y_1]\rightarrow[x_{k_1+1},y_2]\\
&\rightarrow[x_{k_1+1},y_2]\rightarrow[x_{k_1+1},y_2]\rightarrow[x_{k_1+2},y_2]\cdots \rightarrow\cdots
\rightarrow[x_{k_2},y_2]\rightarrow[x_{k_2+1},y_3]\rightarrow\\
&\cdots\rightarrow\cdots\rightarrow[x_{k_N},y_N]
\end{split}
\end{equation}
where $ x_j,j = 1,\cdots, k_N $ is the reasoning state, and $ y_k,k=1,\cdots,N $ is the cognitive states. The reasoning process $ x_j $ happens in working memory, and the cognitive states $ y_k $ is stored in generative memory, when referring to the SOAR cognitive architecture. The stochasticity refers to both the initial distribution of $ x_0 $ and the alternation of cognitive states $ y_k,k=1,\cdots,N $. In fact, reasoning by brute force search is not feasible, there must be cognitive development and cooperation in reasoning. We call this genre of thinking "depth-oriented reasoning". Please check for Figure \ref{Fig20230817_1}.

The time interval between each jump of $ y(t;\omega) $ follows an exponential distribution with parameter $ \lambda $, and it is independent of the location $ y(t;\omega) $ jumps to. A possible probabilistic structure of this model is $$
\Omega = \Omega_{\text{time}}\times\Omega_{\text{space}},\P = \P_{\text{time}}\times\P_{\text{space}}
$$
where $ (\Omega_{\text{time}},\P_{\text{time}}) $ decides time of jumps, and $ (\Omega_{\text{space}},\P_{\text{space}})$ decides the cognitive state where $ y(t;\omega) $ jumps to.

The cognitive space $ \scrY $ is established and under self-construction. $ y(t;\omega) $ facilitates the motion of the thought process $ x(t;\omega) $, and the renewal of $ y(t;\omega) $ is necessary and depends on the current thought $ x(t;\omega) $. Consider a continuous-time system, the control of $ y(t) $ over $ x(t) $ is written in$$
\frac{\d x(t)}{\d t} = \v(x(t),y(t))
$$
The control problem in the language of ordinary differential equations is written as $$
\frac{\d x(t)}{\d t} = \v(x(t),\alpha(t))
$$
where $ \alpha(\cdot) $ is the control process. The Hamilton-Jacobi-Bellman equation describes the spatio-temporal distribution of value. We assume alternation of the cognitive, actually control processes to keep thoughts proceeding; however, it poses a challenge to define the spatio-temporal value. We experience the sense of significant progress in some decisive phases of solving mathematical problems, and might admire some solutions more than others for leveraging classical mathematics and good techniques.

We set the stochastic renewal of the cognitive processes or state exponential($ \lambda $) time interval, and induced by the current thought processes or state $ x\in \scrX $. The alternation of the aligning cognition is grounded on the thoughts in the working memory. For $ x(\cdot;\omega) = x $, the alternation is depicted by the probabilistic language$$
\P(y(t) = y^*\vert x(t) = x, y(t-) =y \text{, }\mathrm{time }\text{ }t\text{ }\mathrm{ makes}\text{ }\mathrm{alternation}) = p(y^*\vert x)= p(y^*\vert \Psi(x))= \psi(x,y^*)
$$
where $ y(t-) $ is the left limit. Here we introduce three notations to describe the transition probability. $ p(y^*\vert x) $ is simply for that $ y^* $, where the cognition alternates to, depends on the current $ x(t) $. $ p(y^*\vert \Psi(x)) $ emphasizes the system contains a decoder of the current thought. $ \psi(x,y^*) $ is short for $  p(y^*\vert \Psi(x)) $, and mostly used in the whole text. We have $$
\sum_{y\in\scrY}\psi(x,y) = 1,\forall x\in \scrX
$$

\begin{hypothesis}
$ \psi $ is continuous with respect to $ x $. Similar thoughts give similar probability to select knowledge.
\end{hypothesis}

In this section and the next, we will discuss the following probabilistic model. We write the mathematical modeling in the following discrete mathematical language:

\begin{equation}
\label{IE_personality_discrete}
\left\{\begin{aligned}
&x(t+dt) = x(t)+\v(x(t),y(t))dt\\
&y(t+dt) = \left\{\begin{aligned}
y(t),\text{ with probability }1-\lambda dt \\
y^*,\text{ with probability }\lambda  \psi(x(t),y^*)dt
\end{aligned} \right. \end{aligned} \right.
\end{equation}
$ \lambda $ is the intensity of the renewal process. $ dt $ may take as $ dt = 1 $, $ dt = 0.1 $. Note the discrete-time form is the true form of computers' experiments. However, for the simplicity of mathematics, we consider the continuous-time form in what follows:

\begin{equation}
\label{IE_personality}
\left\{\begin{aligned}
&\frac{\d x(t)}{\d t} = \v(x(t),y(t))\\
&y(t) \text{ is renewal process of exponential renewal rate } \lambda\\
&\text{At each time of renewal }y(\cdot)\text{ jumps to } y^* \text{ with probability } \psi(x(\cdot),y^*)
\end{aligned} \right.
\end{equation}
Continuous-time form \eqref{IE_personality} provides us with more convenience to analysis than its discrete-time form of \eqref{IE_personality_discrete}. 

$ y(t) $ is a stochastic jump process in the cognitive space $ \scrY $. The memory flow is governed by one information $ y\in\scrY $ $$
\frac{\d x(t)}{\d t} = \v(x(t),y)
$$
for an exponentially distributed period of time. Then the background cognition takes a jump. Then the memory flow $ x(t) $ is governed by a new information $ y^* $ for another exponential time length. Let $ T_1 $ and $ T_2 $ be the two adjacent times of renewal of $ y(t) $. $$
\P(y(T_1) = y) = \psi(x(T_1),y)
$$
and $$
\P(y(T_2) = y^*) = \psi(x(T_2),y^*)
$$
We have that $$
T_2-T_1\sim \mathrm{Exp}(\lambda)
$$ 
$ \mathrm{Exp}(\lambda) $ is probabilistically independent with $ x(\cdot) $ and $ y(\cdot) $. During $ [T_1,T_2] $, $ y $ is the cognition aligned by thoughts $ x(\cdot) $: $$
\frac{\d x(t)}{\d t} = \v(x(t),y),t\in [T_1,T_2]
$$

\begin{hypothesis}
$ y(\cdot) $ takes its initial jump at time $ 0 $ when considering $ \Omega\times\R^+ $. When $ t\gg 0 $, the system is approximated as $ \Omega\times[-\infty,\infty] $.
\end{hypothesis}

The probability density is defined as in primary probability theory:\begin{align*}
&\P(x(t)\in E,y(t)\in F) = \int_E\int_F \rho(x,y;t)\d x \d y\\
&\rho(x,t) = \int_\scrY \rho(x,y;t)\d y \\
&\P(x(t)\in E) = \int_E \rho(x;t)\d x \\
&\int_0^{\infty\text{ }\mathrm{or}\text{ }t}\int_\scrY \rho(x,y,\tau;t)\d y\d \tau = \rho(x;t)
\end{align*}
$ \rho(x,y,\tau;t) $ is the joint Lebesgue density of $ x,y $ and $ \tau $ at the current time $ t $. Consider i.i.d. $ (x(t),y(t))^{(i)};i=1,\cdots,N $, probability density is an approximation of the statistical density when $ N $ is extremely large. Consider $N=$ 1 mol in chemistry, it is around $ 6\times 10^{23} $ amount of particles, which amounts to one spoon in laboratory. Also, when $ N $ goes to infinity, the statistical density of i.i.d. samples goes to the probability density, which is characterised by the classical Bernoulli law of large numbers.

Before we start out our main theorem, two symbols "$ x^* $" and "$ \tau $" are to be defined. Within the time period $ [t-\tau,t] $, $ x(\cdot) $ has taken a determined path and reached $ x $ at time $ t $. We are aware of the ODE governed by information $ y$: $$
\left\{\begin{aligned}
& \frac{\d x(u)}{\d u} = \v(x(u),y) \\
& x(t) = x
\end{aligned}\right.
$$
$ x^*(s;x,y) $ is the $ s $ time before $ t $ of the ODE: $$
x^*(s;x,y) = x(t-s)
$$
or equivalently define the time reversal of the autonomous ODE $$
\left\{\begin{aligned}
& \frac{\d w(u)}{\d u} = -\v(w(u),y) \\
& w(0) = x
\end{aligned}\right.
$$
$ x^*(s;x,y) $ is the time reversal ODE at time s: $$
x^*(s;x,y) = w(s)
$$
We use the symbol $ x^*(s;B,y) $ to denote $$
x^*(s;B,y)\triangleq\{x^{'}:\exists x\in B \text{ s.t. } x^{'} = x^*(s;x,y)\}
$$
time reversal of a set $ B $ by the ODE governed with the information $ y $.

\begin{theorem}\label{20230723Thm1}
All after the above tedious statement, the probability density of $x(t) $, $ \rho(x;t)$ satisfies the following two equations: 
\begin{align}
\label{Eq2_20230723Thm1}
&\frac{\partial }{\partial t}\rho(x,t) + \triangledown_x\cdot\left( \int_0^t\int_\scrY \rho(x,y,\tau;t) \v(x,y) \d y\d \tau\right) = 0\\
\label{Eq1_20230723Thm1}
&\rho(x,y,\tau;t) =\lim\limits_{\epsilon\downarrow 0} \frac{\lambda e^{-\lambda \tau} + (1-e^{-\lambda t})\delta_{t}(\tau)}{|B(x,\epsilon)|} \int_{\scrX}\rho(x^{'};t-\tau)\psi(x^{'},y)\I(x^{'}\in x^*(\tau;B(x,\epsilon),y))\d x^{'}
\end{align}
where $ B(x,\epsilon) $ is the $ \epsilon$-radius ball centred at $ x $, and $ |B(x,\epsilon)| $ is to take Lebesgue measure. $ \rho(x;t) $ and $\rho(x,y,\tau;t) $ depend on each other. $ \lambda $ is the intensity or renewal rate of the jump processes. This is the probabilistic result on flows in the velocity governed by renewal information process, that we shall prove in the next section.
\end{theorem}

\begin{theorem}\label{20230715Thm3.4.1}
In the cases $ t\gg 0 $, the system may be approximated as \begin{align}
\label{IE_personality_PDE_abbr}
&\frac{\partial }{\partial t}\rho(x,t) + \triangledown_x\cdot\left( \int_0^\infty\int_\scrY \rho(x,y,\tau;t) \v(x,y) \d y\d \tau\right) = 0\\
\label{Eq_rho_x_y_tau_t_abbr}
&\rho(x,y,\tau;t) =\lim\limits_{\epsilon\downarrow 0} \frac{\lambda e^{-\lambda \tau}}{|B(x,\epsilon)|} \int_{\scrX}\rho(x^{'};t-\tau)\psi(x^{'},y)\I(x^{'}\in x^*(\tau;B(x,\epsilon),y))\d x^{'}
\end{align}
by assuming time in $ [-\infty,\infty] $. $ \lambda $ is the intensity of the jump processes. This is the probabilistic result on flows in the velocity governed by renewal information process, that we shall prove in the next section.
\end{theorem}

\section{Proof of the Theorem in the Previous Section}
This section is dedicated to providing proof of Theorem \ref{20230715Thm3.4.1}.

\begin{remark}[Notation Convention]
We use a specific symbol in the text of this section. We frequently use the symbol $ \partial E + (\v\cdot\mathbf{n}) \mathbf{n}\Delta t $ to represent a narrow boundary of $ E $, to be defined as follows: $$
\partial E + (\v\cdot\mathbf{n}) \mathbf{n}\Delta t  \triangleq \{x\in\scrX:\exists x^*\in\partial E, \exists 0\leqslant s\leqslant \Delta t , x = x^* + \left(\v(x^*)\cdot\mathbf{n}s\right) \mathbf{n}\}
$$
Here $ \mathbf{n} $ as usual denotes the unit outward normal vector of $ \partial E $. The idea of proposing this symbol is that the very narrow band of $ \partial E $ is expanded by shifting velocity field $ \v $ over an instant time period $\Delta t$ alongside $ \partial E $ the boundary of $ E $. In other words, imagine we have a group of particles starting at $ \partial E $ moving along $ v $, then $ \partial E + (\v\cdot\mathbf{n}) \mathbf{n}\Delta t $ represents their trajectories within instant time period $ \Delta t $. In fact, considering the motion quantity of order $ \Delta t ^2 $, the narrow band can be written as $$
\partial E + (\v\cdot\mathbf{n}) \mathbf{n}\Delta t + (\left[\left(\partial_t + \v\cdot\partial_x\right)\v\right]\cdot\mathbf{n}) \mathbf{n}\Delta t^2 + o(\Delta t^2)
$$
Given the architecture of $$
\frac{\partial A}{\partial t}\Delta t + B\Delta t + o(\Delta t) = 0
$$
we only need to consider the band of order $ \Delta t $ and ignore $ o(\Delta t) $.\qedsymbol
\end{remark}

Let us first prove \eqref{Eq1_20230723Thm1} and \eqref{Eq_rho_x_y_tau_t_abbr} by the following Lemma \ref{Lem3.4.2} and Proof \ref{Prof20230726}:

\begin{lemma}\label{Lem3.4.2}
$ \tau(t;\omega) $ is the stochastic process referring to the elapsed time since the renewal cognitive stream $ y(\cdot;\omega) $ made its last jump until time $ t $: $$
y(s;\omega) = y(t-\tau(t;\omega);\omega),\forall s\in [t-\tau(t;\omega),t]
$$
Given the current time $ t $, the stochastic processes $ x(\cdot;\omega), y(\cdot;\omega), \tau(\cdot;\omega) $ degenerate to random variables $ x(t;\omega), y(t;\omega), \tau(t;\omega) $. Let $ \rho(x,y,\tau;t) $ denotes their probability density, and let $ \rho(x;t) $ denotes the probability density of $ x(t;\omega) $: $$
\rho(x;t) = \iint \rho(x,y,\tau;t)\d y\d \tau
$$

Please consider the motion starting at time $ 0 $. The initial probability density of $ x(0;\omega) $ is $ \rho(x;0)
$ and $ y(\cdot) $ took its initial jump at time $ 0 $. The joint probability density at time $ t $, of the reasoning memory flow $ x(t) $, the informed cognitive flow $ y(t) $, the elapsed time $ y(\cdot) $ took a jump $ \tau $, $ \rho(x,y,\tau;t) $, is given by 
$$
\rho(x,y,\tau;t) =\lim\limits_{\epsilon\downarrow 0} \frac{\lambda e^{-\lambda \tau} + (1-e^{-\lambda t})\delta_{t}(\tau)}{|B(x,\epsilon)|} \int_{\scrX}\rho(x^{'};t-\tau)p(y\vert\Psi(x^{'}))\I(x^{'}\in x^*(\tau;B(x,\epsilon),y))\d x^{'}
$$
where $ p(y\vert\Psi(x^{'})) $ is abbreviated to $ \psi(x^{'},y) $. If further consider motion in full time period of $ (-\infty,\infty) $, the joint probability density at time $ t $ is given by 
$$
\rho(x,y,\tau;t) =\lim\limits_{\epsilon\downarrow 0} \frac{\lambda e^{-\lambda \tau}}{|B(x,\epsilon)|} \int_{\scrX}\rho(x^{'};t-\tau)p(y\vert\Psi(x^{'}))\I(x^{'}\in x^*(\tau;B(x,\epsilon),y))\d x^{'}
$$
These two are \eqref{Eq1_20230723Thm1} and \eqref{Eq_rho_x_y_tau_t_abbr}, and it is the same as rewriting \eqref{Eq1_20230723Thm1} and \eqref{Eq_rho_x_y_tau_t_abbr} as \begin{equation}
\begin{split}
&\rho(x,y,\tau;t)|B(x,\epsilon)|\Delta \tau  + o(\epsilon) + o(\Delta\tau)\\
=& \rho(x^*(\tau;x,y);t-\tau)\psi(x^*(\tau;x,y),y)\Big|\{x^{'}:x^{'}\in x^*(\tau;B(x,\epsilon),y)\}\Big|\Big(\lambda e^{-\lambda \tau} + (1-e^{-\lambda t})\delta_{t}(\tau)\Big)\Delta\tau \\
&\rho(x,y,\tau;t)|B(x,\epsilon)|\Delta \tau  + o(\epsilon) + o(\Delta\tau)\\
=& \rho(x^*(\tau;x,y);t-\tau)\psi(x^*(\tau;x,y),y)\Big|\{x^{'}:x^{'}\in x^*(\tau;B(x,\epsilon),y)\}\Big|\lambda e^{-\lambda \tau}\Delta\tau 
\end{split}
\end{equation}
where $|\cdot| $ denotes taking the area (Lebesgue measure) of the set. In the beginning time $ 0 $ of \eqref{Eq1_20230723Thm1}, the distribution function of $ \tau $ exhibits a jump with a magnitude of $ (1-e^{-\lambda t}) $, resulting in the density function having a Dirac delta function. In the case $ t\gg \lambda $, the two cases above, \eqref{Eq1_20230723Thm1} and \eqref{Eq_rho_x_y_tau_t_abbr}, are similar.\qedsymbol
\end{lemma}

\begin{prof}[Proof of Lemma \ref{Lem3.4.2}]\label{Prof20230726}
The idea of proof is quite straightforward: Given $ \tau(t) $ and $ y(t) $, $ x(t) $ can be with no stochasticity reversed back to $\tau $-previous time, the moment $ x(t-\tau) $ occurs. For every particle $ \omega $, $ x(t;\omega) $ has unique reversal time $ t-\tau(t;\omega) $ and reversal state $ x(t-\tau;\omega) $, which is the fact we are using to derive our result. Consider spatial domain $$
B\times C\subset \scrX\times\scrY
$$
and temporal domain $ [t_1,t_2] $, $$
[t_1,t_2]\subset [0,t]
$$
for deriving \eqref{Eq1_20230723Thm1} as well as  $$
[t_1,t_2]\subset [0,\infty)
$$
for deriving \eqref{Eq_rho_x_y_tau_t_abbr}. We are expressing the probability density $ \rho(x,y,\tau;t) $ with a unique reversal state $$
\Big(x(t-\tau(t;\omega);\omega), y(t;\omega)\Big) = \Big( x(t-\tau(t;\omega);\omega), y(t-\tau(t;\omega);\omega) \Big)
$$or written simply as $$
\Big(x(t-\tau), y(t)\Big) = \Big( x(t-\tau), y(t-\tau) \Big)
$$for every particle $ \omega $.

By the unique representation of reversal state, we have that $$
\P(x(t)\in B, y(t)\in C, \tau\in[t_1,t_2] )  =\P(x(t-\tau)\in x^*(\tau;B,y(t)),y(t)\in C, \tau\in[t_1,t_2])
$$
where $ x^* $ as defined in the last section is the time reversal of the ODE $$
\left\{\begin{aligned}
&\frac{\d x(t)}{\d t} = \v(x(t),y)\\
&x(t) = u
\end{aligned}\right.
$$
with $$
x^*(s;u,y) = x(t-s)
$$
$ s $ denotes reversing to $ s $-previous time, $ u $ is the current position of $ x(t) $, and $ y $ is the parameter of motion velocity. We slightly abuse the symbol $ x^* $: $$
x^*(s;B,y) = \{u^{'}:\exists u \in B\text{ s.t. }u^{'} = x^*(s;u,y) \}
$$
is the time reversal of set $ B $.

The right hand side \begin{align*} 
&\P(x(t-\tau)\in x^*(\tau;B,y(t)),y(t)\in C, \tau\in[t_1,t_2])\\
=&\int_{[t_1,t_2]} \lambda \exp(-\lambda s)\P(x(t-\tau)\in x^*(\tau;B,y(t)),y(t)\in C\vert \tau = s)\d s\\
=&\int_{[t_1,t_2]} \lambda \exp(-\lambda s)\P(x(t-s)\in x^*(s;B,y(t)),y(t)\in C\vert \tau = s)\d s
\end{align*}
The jumps occur independently of particles both in terms of time and space, the event $
\{\tau = s\}
$
is probabilistically independent with the spatial location $
\{x(t-s)\in x^*(\tau;B,y(t)),y(t)\in C\}
$
To write in full mathematical language: \begin{align*}
&\P(x(t-s)\in x^*(s;B,y(t)),y(t)\in C\vert \tau = s)\\
=&\lim_{\Delta s\downarrow 0} \frac{\P(x(t-s)\in x^*(s;B,y(t)),y(t)\in C, \tau \in [s,s+\Delta s])}{\P(\tau \in [s-\Delta s,s])}\\
=&\lim_{\Delta s\downarrow 0} \frac{\P(x(t-s)\in x^*(s;B,y(t)),y(t-s )\in C, \tau \in [s,s+\Delta s])}{\P(\tau \in [s-\Delta s,s])}
\end{align*}
This is because $ y(\cdot) $ took its last jump at $ t-s $. We proceed \begin{align*}
&\lim_{\Delta s\downarrow 0} \frac{\P(x(t-s)\in x^*(s;B,y(t)),y(t-s )\in C, \tau \in [s-\Delta s,s])}{\P(\tau \in [s-\Delta s,s])}\\
=&\lim_{\Delta s\downarrow 0} \frac{\P(x(t-s)\in x^*(s;B,y(t)),y(t-s )\in C, \tau \in [s-\Delta s,s])}{\P(x(t-s)\in \scrX,y(t-s )\in \scrY,\tau \in [s-\Delta s,s])}\\
=&\lim_{\Delta s\downarrow 0} \frac{\iint_{\scrX\times\scrY}\rho(x^{'};t-s)\psi(x^{'},y)\I(x^{'}\in x^*(s;B,y), y\in C) \d y\d x^{'} (\lambda\Delta s + o(\Delta s))\exp(-\lambda s)}{(\lambda\Delta s + o(\Delta s))\exp(-\lambda s)}\\
=&\iint_{\scrX\times\scrY}\rho(x^{'};t-s)\psi(x^{'},y)\I(x^{'}\in x^*(s;B,y), y\in C) \d y\d x^{'} 
\end{align*}
Now we have that \begin{align*}
&\P(x(t)\in B, y(t)\in C, \tau\in[t_1,t_2] )\\
=&\int_{[t_1,t_2]} \lambda \exp(-\lambda s)\int_C\int_{\scrX}\rho(x^{'};t-s)\psi(x^{'},y)\I(x^{'}\in x^*(s;B,y))\d x^{'}\d y\d s
\end{align*}
By taking $ B = B(x,\epsilon)$ ,$ C= B(y,\epsilon) $, and let $ \epsilon \downarrow 0 $ and $ t_2\downarrow t_1 $ we derive the result of Lemma \ref{Lem3.4.2}.\qedsymbol
\end{prof}

\begin{prof}[Another Proof of Lemma \ref{Lem3.4.2}]
Here give another proof for the lemma. The technique is writing the integral as limit of the summation: We write $$
\P(x(t)\in B, y(t)\in C, \tau\in[t_1,t_2] )
$$
to the summation $$
\sum\limits_{\Delta s\in[t_1,t_2]} \P(x(t)\in B, y(t)\in C, \tau\in[s,s+\Delta s] )
$$
We proceed \begin{align*}
=\sum\limits_{\Delta s\in[t_1,t_2]} &\P(x(t-s)\in x^*(s;B,y(t)), y(t)\in C, \tau\in[s,s+\Delta  s] )\\
=\sum\limits_{\Delta s\in[t_1,t_2]} &\P(x(t-s)\in x^*(s;B,y(t)), y(t)\in C, y(\cdot)\text{took a jump at time }t_2-s,\\
 &y(\cdot)\text{ would take no jumps at }[t_2-s,t_2] )
\end{align*}
The possible particles that enters $ B\times C\subset\scrX\times\scrY $ is distributed $ x^*(\tau;B,y))\times C $ after the jump of $ y(\cdot) $. The probability to take a jump in $ \scrY $, for every particle, including them for sure, is $ \lambda\Delta s $, independent with the probability of no longer taking jumps $ \exp(-\lambda s) $. We proceed $$
\iint_{\scrX\times\scrY}\rho(x^{'};t-s)\psi(x^{'},y)\I(x^{'}\in x^*(\tau;B,y))\I(y\in C)\d y\d x^{'}\exp(-\lambda s) \lambda\Delta s 
$$
Rewrite summation to integral by taking $ \Delta s \downarrow 0 $: $$
=\int_{[t_1,t_2]} \lambda \exp(-\lambda s)\iint_{\scrX\times\scrY}\rho(x^{'};t-s)\psi(x^{'},y)\I(x^{'}\in x^*(\tau;B,y))\I(y\in C) \d y\d x^{'}\d s
$$
By taking $ B = B(x,\epsilon)$ ,$ C= B(y,\epsilon) $, and let $ \epsilon \downarrow 0 $ and $ t_2\downarrow t_1 $ we derive the result of Lemma \ref{Lem3.4.2}.\qedsymbol
\end{prof}

Both methods are intended to illustrate the fact that \begin{align*}
\P(x(t-s)\in x^*(s;B,y(t)),&y(t)\in C\vert \tau = s) \\
&= \iint_{\scrX\times\scrY}\rho(x^{'};t-s)\psi(x^{'},y)\I(x^{'}\in x^*(\tau;B,y), y\in C) \d y\d x^{'}
\end{align*}
Finally, let us turn to prove \eqref{Eq2_20230723Thm1} and  \eqref{IE_personality_PDE_abbr}:

\begin{figure}[htb]
\centering
\includegraphics[width = 0.8\textwidth]{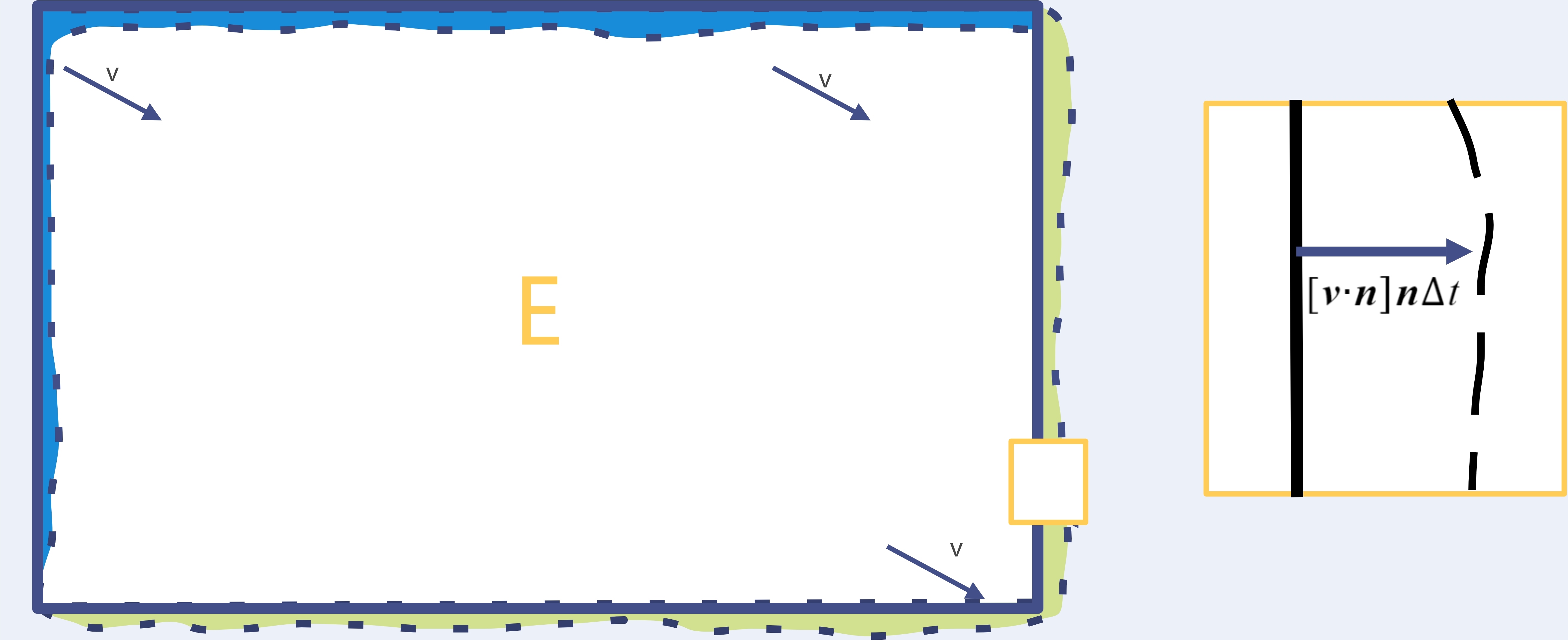}
\caption{Illustration of Case (i)}
\label{Fig20230811}
\end{figure}

\begin{remark}[A remark for Proof \ref{prof20230730}]
Consider a domain (closed, connected, and bounded subset with smooth boundary) $ E \subset\scrX $ and instantaneous time interval $ [t,t+\Delta t] $. The increment of $ x(\cdot;\omega) $ in $ E $ during $ [t,t+\Delta t] $ is $$
\int_E\left(\rho(x;t+\Delta t) - \rho(x;t)\right)\d x = \left(\int_E \frac{\partial \rho(x;t)}{\partial t}\d x\right)\Delta t + o(\Delta t)
$$
which is the amount of $ x(\cdot;\omega) $ that goes in $ E $ minus the amount of $ x(\cdot;\omega) $ that goes out of $ E $. This increment is categorized into three groups: \begin{enumerate}[label=(\roman*)]
\item $ x(\cdot;\omega) $ with the accompanying $ y(\cdot;\omega) $ making no jumps in $ \scrY $ during $ [t,t+\Delta t] $.
\item $ x(\cdot;\omega) $ with the accompanying $ y(\cdot;\omega) $ making one jump in $ \scrY $ during $ [t,t+\Delta t] $.
\item $ x(\cdot;\omega) $ with the accompanying $ y(\cdot;\omega) $ making more than two jumps in $ \scrY $ during $ [t,t+\Delta t] $.
\end{enumerate} We quantify the architecture: $$
\left(\int_E \frac{\partial \rho(x;t)}{\partial t}\d x\right)\Delta t + o(\Delta t) = \text{ (i) + (ii) + (iii)}
$$

With regard to $ (i) $, our goal is to derive an expression of \begin{align*}
\text{Case (i)} =& -\E\Big[\Big(\I[\v(x(t),y(t))\cdot\mathbf{n}>0]-\I[\v(x(t),y(t))\cdot\mathbf{n}<0]\Big)\\
&\I\Big(x(t)\in\partial E + (\v(\cdot,y(t))\cdot\mathbf{n})\mathbf{n}\Delta t \Big)\Big] + o(\Delta t)
\end{align*} 
Here we apply a symbol of $$
\partial E + (\v(\cdot,y(t))\cdot\mathbf{n})\mathbf{n}\Delta t
$$
to denote the narrow band along the boundary $ \partial E $:
$$
\partial E + (\v(\cdot,y(t))\cdot\mathbf{n})\mathbf{n}\Delta t \triangleq \{x\in\scrX:\exists x^*\in\partial E, \exists 0\leqslant s\leqslant \Delta t \text{ s.t. } x = x^* + (\v(x^*,y(t))\cdot\mathbf{n})\mathbf{n} s\}
$$
We need $$
\Big(\I(\v(x(t),y(t))\cdot\mathbf{n}>0)-\I(\v(x(t),y(t))\cdot\mathbf{n}<0)\Big)
$$
before the boundary band. When $ \v\cdot\mathbf{n} $ is greater than zero, particles flow out; while when $ \v\cdot\mathbf{n} $ is less than zero, particles flow in. Notice $ y(t) $ is random variable.

Conditioning on $ y(\cdot;\omega) = y $, particles flowing into $ E $ are distributed $$
\left\{\partial E + (\v(\cdot,y)\cdot\mathbf{n})\mathbf{n}\Delta t + o(\Delta t) \right\}\cap\{x^{'}:x^{'} = x + \v(x,y)s, x\in \partial E, 0\leqslant s\leqslant \Delta t, \v(x,y)\cdot\mathbf{n}<0\}
$$particles flowing out of $ E $ are distributed $$
\left\{\partial E + (\v(\cdot,y)\cdot\mathbf{n})\mathbf{n}\Delta t + o(\Delta t) \right\}\cap\{x^{'}:x^{'} = x + \v(x,y)s, x\in \partial E, 0\leqslant s\leqslant \Delta t, \v(x,y)\cdot\mathbf{n}>0\}
$$
where $ \mathbf{n} $ is the outward normal unit vector of $ \partial E $ the boundary of $ E $. The scenario is illustrated by Figure \ref{Fig20230811}. By the Gauss-Green theorem in calculus, the increment is $$
-\int_{\partial E}\rho(x\vert y;t)\v(x,y)\cdot\mathbf{n}\d x\Delta t +o(\Delta t)=-\int_E \triangledown_x\cdot[\rho(x\vert y;t)\v(x,y)]\d x\Delta t +o(\Delta t)
$$
Taking average on $ y $, the increment is $$
(i) = \int_\scrY\int_E -\triangledown_x\cdot[\rho(x\vert y;t)\v(x,y)]\d x\rho(y;t)\d y\Delta t +o(\Delta t) = -\int_E\int_\scrY \triangledown_x\cdot[\v(x,y)\rho(x,y;t)]\d y\d x\Delta t +o(\Delta t)
$$

By replacing \begin{align*}
&\rho(x,y;t) = \int_0^t \rho(x,y,\tau; t)\d\tau \\
\text{   or   } &\rho(x,y;t) = \int_0^\infty \rho(x,y,\tau; t)\d\tau
\end{align*}
we conclude the result \begin{align*}
\int_{[t,t+\Delta t]}\int_E \frac{\partial}{\partial t}\rho(x,s)\d x\d s
&= -\int_E \triangledown_x\cdot\left( \int_0^{\infty \text{ }\mathrm{or}\text{ } t}\int_\scrY \rho(x,y,\tau;t) \v(x,y) \d y\d \tau\right) \d x \Delta t +o(\Delta t)\\
&+ \text{ Case Two } + \text{ Case Three }
\end{align*}
$ \Delta t $ is an infinitesimally small time interval, and the remaining task is to prove that Case (ii) and Case (iii) in the last equation both yield $ o(\Delta t) $. After showing that Case (ii) and Case (iii) bring $ o(\Delta t) $, we are secure to say\begin{align*}
\frac{\partial }{\partial t}\rho(x,t) + \triangledown_x\cdot\left( \int_0^{\infty \text{ }\mathrm{or}\text{ } t}\int_\scrY \rho(x,y,\tau;t) \v(x,y) \d y\d \tau\right)= 0
\end{align*}
with Case (ii) and Case (iii) vanished in the mathematics of temporally first-order partial differential equation.

With regard to $ (ii) $, the probability $ y(\cdot;\omega) $ to make one jump in $ [t,t+\Delta t] $ is $
\lambda\Delta t
$
and the spatial location $ x(\cdot;\omega) $ that enter or exit $ E $ is $$
\partial E + V\Delta t
$$
the $ \Delta t $-breadth boundary of $ E $, where $ V $ is a certain vector. The spatial location is independent of the jumps, and thus $$
(ii) \sim\lambda\Delta t\cdot V\Delta t = o(\Delta t)
$$
However, providing a strict proof of this requires a detailed and intricate discussion.

With regard to $ (iii) $, the probability of more than two jumps occurring is $ o(\Delta t) $, the affected particles in the increment of $ E $ is the same magnitude as $ \Delta t $, thus the increment $$
(iii)\sim\Delta t\cdot o(\Delta t)
$$\qedsymbol
\end{remark}

\begin{prof}[Proof of \eqref{Eq2_20230723Thm1} and \eqref{IE_personality_PDE_abbr}]\label{prof20230730}
Consider $ E \subset\scrX $ and instantaneous time interval $ [t,t+\Delta t] $. There are three cases to consider in division. Case One: $ y(t) $ does not jump within the time interval $ [t,t+\Delta t] $. Case Two: $ y(t) $ has one jump within the time interval $ [t,t+\Delta t] $.  Case Three: $ y(t) $ takes more than two jumps within the time interval $ [t,t+\Delta t]$. We are going to show that Case Two and Case Three can be eliminated as their quantities are $ o(\Delta t) $. The proof of the theorem is to fill in the following structure: $$
\left(\int_E \frac{\partial \rho(x;t)}{\partial t}\d x\right)\Delta t + o(\Delta t) = \text{ (Case One) + (Case Two) + (Case Three)}
$$
The formulas of the law of iterated expectations and the law of total probability are unified as one in advanced probability theory.

We first consider Case One: $ y(t) $ does not jump within the time interval $ [t,t+\Delta t] $. As illustrated in Figure \ref{Fig20230811}, particles that enter or exit $ E $ during the time interval $ [t,t\Delta t] $ form a narrow band along $ \partial E $ the boundary of region $ E $: $$
\partial E + [\v(*,y(t))\cdot\mathbf{n}]\mathbf{n}\Delta t + o(\Delta t),*\in\partial E
$$
When $ \v(*,y(t))\cdot\mathbf{n} <0 $ particles enter $ E $, while when $ \v(*,y(t))\cdot\mathbf{n} >0 $ particles enter $ E $, $ * $ is the element along $ \partial E $ which points to $ x(t+\Delta t) $. We have a fast perception that \begin{align*}
\text{Case One} =& -\E\Big[\Big(\I[\v(x(t),y(t))\cdot\mathbf{n}>0]-\I[\v(x(t),y(t))\cdot\mathbf{n}<0]\Big)\\
&\I\Big(x(t)\in\partial E + (\v(\cdot,y(t))\cdot\mathbf{n})\mathbf{n}\Delta t \Big)\Big] + o(\Delta t)
\end{align*} 
however, the strict proof of this requires writing out all the details.

To be precise in expression, particles that enter $ E $ during $ [t,t+\Delta t] $ are \begin{equation}\label{Eq20230811_1}
\{\omega:\exists u_{\omega,t}\in\partial E, x(t+\Delta t;\omega)\in\partial E + [\v(u_{\omega,t},y(t))\cdot\mathbf{n}]\mathbf{n}\Delta t + o(\Delta t),\v(u_{\omega,t},y(t))\cdot\mathbf{n}<0\}
\end{equation}
Particles that exit $ E $ during $ [t,t+\Delta t] $ are \begin{equation}\label{Eq20230811_2}
\{\omega:\exists v_{\omega,t}\in\partial E, x(t+\Delta t;\omega)\in\partial E + [\v(v_{\omega,t},y(t))\cdot\mathbf{n}]\mathbf{n}\Delta t + o(\Delta t),\v(v_{\omega,t},y(t))\cdot\mathbf{n}>0\}
\end{equation}
The $\{ u_{\omega,t}\} $ and $\{v_{\omega,t}\} $ in \eqref{Eq20230811_1} and \eqref{Eq20230811_2} together are the boundary of $ E $, always denoted by $ \partial E $ in this article. They are functions of $ x(t;\omega) $ and $ t $, later denoted by $\{ u_{x,t}\} $ and $\{v_{x,t}\} $ in the discussion of probability density. $\{ u_{x,t}\} $ are the points on the boundary where particles go in, while $\{v_{x,t}\} $ are the positions particles flow out of $ E $.

We proceed \begin{align}
&\notag \P\{\omega:\exists u_{\omega,t}\in\partial E, x(t+\Delta t;\omega)\in\partial E + [\v(u_{\omega,t},y(t))\cdot\mathbf{n}]\mathbf{n}\Delta t + o(\Delta t),\v(u_{\omega,t},y(t))\cdot\mathbf{n}<0\}\\
&\notag -\P\{\omega:\exists v_{\omega,t}\in\partial E, x(t+\Delta t;\omega)\in\partial E + [\v(v_{\omega,t},y(t))\cdot\mathbf{n}]\mathbf{n}\Delta t + o(\Delta t),\v(v_{\omega,t},y(t))\cdot\mathbf{n}>0\}\\
=&\notag \iint_{\scrX\times\scrY} \I[x\in \partial E + [\v(u_{x,t},y)\cdot\mathbf{n}]\mathbf{n}\Delta t + o(\Delta t)]\rho(x,y;t+\Delta t)\d y\d x \\
&\label{Eq20230813_1}- \iint_{\scrX\times\scrY} \I[x\in \partial E + [\v(v_{x,t},y)\cdot\mathbf{n}]\mathbf{n}\Delta t + o(\Delta t)]\rho(x,y;t+\Delta t)\d y\d x\\
=&\notag \iint_{\scrX\times\scrY} \I[x\in \partial E + [\v(u_{x,t},y)\cdot\mathbf{n}]\mathbf{n}\Delta t ]\rho(x,y;t+\Delta t)\d y\d x \\
&\label{Eq20230813_2}- \iint_{\scrX\times\scrY} \I[x\in \partial E + [\v(v_{x,t},y)\cdot\mathbf{n}]\mathbf{n}\Delta t ]\rho(x,y;t+\Delta t)\d y\d x + o(\Delta t)\\
=&\notag \iint_{\scrX\times\scrY} \I[x\in \partial E + [\v(u_{x,t},y)\cdot\mathbf{n}]\mathbf{n}\Delta t ]\rho(x,y;t)\d y\d x \\
&\label{Eq20230813_3}- \iint_{\scrX\times\scrY} \I[x\in \partial E + [\v(v_{x,t},y)\cdot\mathbf{n}]\mathbf{n}\Delta t ]\rho(x,y;t)\d y\d x + o(\Delta t)+ o(\Delta t)\\
=&\notag \iint_{\scrX\times\scrY} \I[x\in \partial E + [\v(u_{x,t},y)\cdot\mathbf{n}]\mathbf{n}\Delta t ]\rho(u_{x,t},y;t)\d y\d x \\
&\label{Eq20230813_4}- \iint_{\scrX\times\scrY} \I[x\in \partial E + [\v(v_{x,t},y)\cdot\mathbf{n}]\mathbf{n}\Delta t ]\rho(v_{x,t},y;t)\d y\d x + o(\Delta t)+ o(\Delta t)\\
=&\label{Eq20230813_5}- \int_{\partial E} \int_\scrY \rho(s,y;t)[\v(s,y)\cdot\mathbf{n}]\Delta t\d y\d s + o(\Delta t)+ o(\Delta t)+ o(\Delta t)\\
=&\label{Eq20230813_6}- \int_{E} \int_\scrY \triangledown_x\cdot[\rho(x,y;t)\v(x,y)]\d y\d x\Delta t + o(\Delta t)
\end{align}
\eqref{Eq20230813_1} is because $ y(t) = y(t+\Delta t) $ in Case One; $ u_{x,t} $ and $ v_{x,y} $ points to $ x $ from $ \partial E $. \eqref{Eq20230813_2} is because the narrow band along $ \partial E $ is approximated by $ \partial E + [\v(u_{x,t},y)\cdot\mathbf{n}]\mathbf{n}\Delta t \cup \partial E + [\v(v_{x,t},y)\cdot\mathbf{n}]\mathbf{n}\Delta t $ yielding additional $ o(\Delta t) $. \eqref{Eq20230813_3} is because $ \rho(x,y;t+\Delta t) $ differ from $ \rho(x,y;t) $ with a difference of $ o(\Delta t) $. \eqref{Eq20230813_4} is because $ \rho(x,y;t) $ differ from $ \rho(u_{x,t},y;t) $ and $ \rho(v_{x,t},y;t) $ yielding $ o(\Delta t) $. \eqref{Eq20230813_5} is because there adds additional minus sign for particles that enter $ E $. \eqref{Eq20230813_6} is the Gauss-Green theorem in calculus.

We have shown that$$
\left(\int_E \frac{\partial \rho(x;t)}{\partial t}\d x\right)\Delta t + o(\Delta t) = - \int_{E} \int_\scrY \triangledown_x\cdot[\rho(x,y;t)\v(x,y)]\d y\d x\Delta t + o(\Delta t) \text{+ (Case Two) + (Case Three)}
$$
it remains to show that Case Two and Case Three above are all $ o(\Delta t) $ to finish the proof.

Readers familiar with the theory of probability easily rule out Case Three as $ o(\Delta t) $: The probability of making two jumps within $ [t,t+\Delta t] $ is $ o(\Delta t) $, and the total spatial measure of particles is $ 1 $. Here the notation of $ \Omega_{\mathrm{space}}\times\Omega_{\mathrm{time}} $ makes the strict illustration. The set of particles making more than two jumps and entering or exiting $ E $ is denoted as $$
E_{2}\subset \Omega = \Omega_{\mathrm{space}}\times\Omega_{\mathrm{time}} 
$$
Then \begin{align*}
\P(E_2)&\leqslant \P[\Omega_{\mathrm{space}}\times \mathrm{Proj}(F,\Omega_{\mathrm{time}})]\\
&=\P_{\mathrm{time}}[\mathrm{Proj}(F,\Omega_{\mathrm{time}})]\\
&\leqslant \P_{\mathrm{time}}[\text{all the particles that take more than two jumps within }[t,t+\Delta t] ]\\
&=o(\Delta t)
\end{align*}
$ \mathrm{Proj}(F,\Omega_{\mathrm{time}}) $ means the projection of $ E_2 $ to its marginal dimension $ \Omega_{\mathrm{time}} $. Since the probability of a Poisson renewal process taking more than two renewals in $ [t,t+\Delta t] $ is $ o(\Delta t) $, even all the particles are gathered around $ \partial E $, the result is still $ o(\Delta t) $ as you may guess; however, to prove Case Two $o(\Delta t) $ is difficult and rather involved.

The proof of Case Two is to analyze the infinite partitioning of the instantaneous $ \Delta t $. Consider an partitioning interval of length $\d s $: $[t+s,t+s+\d s]\subset [t,t+\Delta t] $. Given the knowledge $ y(\cdot;\omega) $ taking one jump within $ [t,t+\Delta t] $ with probability $ \lambda\Delta t + o(\Delta t) $ for Poisson renewal processes, the jump moment lies in $ [t+s,t+s+\d s] $ with conditional probability $$
\frac{\d s}{\Delta t}
$$
We proceed to analyze $$
\E\big[\v(x,y(t+s)\big\vert x(t+s) = x\big]\qquad\mathrm{and}\qquad \E\big[\v(x,y(t+s+\d s)\big\vert x(t+s) = x\big]
$$
the conditionally expected velocity. Conditioning on $ x(t+s) = x $, the expected velocity before the jump of $ y(\cdot) $, at time $ t+s $ is $$
\v_A(x) = \E\big[\v(x,y(t+s)\big\vert x(t+s) = x\big] = \int_\scrY \rho(y\vert x;t) \v(x,y) \d y
$$ 
where $$
\rho(y,\tau\vert x;t) = \rho(x,y,\tau;t)/\rho(x;t)\qquad\mathrm{and}\qquad \rho(y\vert x;t) = \int\rho(y,\tau\vert x;t)\d\tau
$$

\begin{proposition}\label{Prop20230815_2}
Conditioning on $ x(t+s) = x $, the expected velocity after the jump, at time $ t+s+\d s $ is $$
\v_B(x) = \E\big[\v(x,y(t+s+\d s)\big\vert x(t+s) = x\big] = \int_\scrY \v(x,y)\psi(x,y)\d y
$$ 
We give the analytic form of $ \v_B(x) $, which depends on $ x $ and $ s $, for clarity of the text; however, this analytic form is not necessary for later work.
\end{proposition}
\begin{prof}[Proof of Proposition \ref{Prop20230815_2}]
Let $ t+s+\theta\d s $ be the exact time of jump. $ \theta $ is a random variable of uniform distribution on $ [0,1] $. We have \begin{align*}
&\int_0^1\int_\scrY \left[\rho(x^{'};t+s+\theta\d s)\left|\frac{\partial x^{'}}{\partial x}\right|\psi(x^{'},y)\v(x^{'},y)\right]\d y\d\theta \Bigg\vert_{x^{'} =  x+\v_A \theta\d s + o(\d s)}\\
=& \rho(x;t+s)\E\big[\v(x,y(t+s+\d s)\big\vert x(t+s) = x\big]
\end{align*}
$ \left|\frac{\partial x^{'}}{\partial x}\right| $ is the diffusion effect of $ \v_A $ and is approximated as $ 1 + O(\d s) $. Because $ \rho $ is continuous and $ \psi $ is continuous with respect to $ x^{'} $, the left hand side is $$
\rho(x;t)\int_\scrY \v(x,y)\psi(x,y)\d y + O(\d s)
$$
Since $ O(\d s) $ vanishes in the later integral of $ s $, we omit writing the term $ o(\d s) $ in $$
\v_B(x) = \int_\scrY \v(x,y)\psi(x,y)\d y
$$\qedsymbol
\end{prof}

\begin{proposition}\label{Prop20230815}
Consider the particles $ \{\omega\}\subset\Omega $ that take one jump in $ \scrY $ during $[t+s,t+s+\d s]\subset[t,t+\Delta t] $. The spatial measure on the space of $ \scrX $ of these particles is $$
(\Delta t - s)(-\v_B + \v_A)\cdot\mathbf{n}
$$
encompassing four possible scenarios in the proof.
\end{proposition}
\begin{prof}[Proof of Proposition \ref{Prop20230815}]
Let us consider the following four cases:
\begin{enumerate}
\item $ \v_A $ is pointing outward of $ E $ and $ \v_B $ is pointing inward of $ E $.

Particles flowing in are spatially distributed $$ \partial E + (-\v_B\cdot\mathbf{n})\mathbf{n}(\Delta t - s) $$ 
these particles should not flow in if not happens the jump of $ y(\cdot) $.

Particles supposed to flow out are spatially distributed $$ \partial E - (\v_A\cdot\mathbf{n})\mathbf{n}(\Delta t - s) $$
these particles will not flow out because of the jump.

\item $ \v_A $ is pointing inward of $ E $ and $ \v_B $ is pointing outward of $ E $.

Particles supposed to enter $ E $ are distributed $$ \partial E + (-\v_A\cdot\mathbf{n})\mathbf{n}(\Delta t - s) $$ 
at time $ t+s $; however, the jump interrupts them from flowing in.

Particles that change their direction because of the alternation of $ y(\cdot) $ and flow out are spatially distributed $$ \partial E - (\v_B\cdot\mathbf{n})\mathbf{n}(\Delta t - s) $$
at time $ t+s $.

Notice that both Case 1 and Case 2 reveal a result of change of $ E $ by the once jump of the cognitive that induces velocity changes: $$
(-\v_B + \v_A)\cdot\mathbf{n}(\Delta t - s)
$$

\item $ \v_A $ is pointing outward of $ E $ and $ \v_B $ is pointing outward of $ E $. When $ \v_B $ is longer than $ \v_A $, $$
\{\partial E - (\v_B)\cdot\mathbf{n}\mathbf{n} (\Delta t-s)\} - \{\partial E - (\v_A)\cdot\mathbf{n}\mathbf{n} (\Delta t-s)\}
$$ 
distributes the particles that additionally turn to flow out because of the cognitive alternation. $ - $ in the middle is the set minus. When $ \v_B $ is shorter than $ \v_A $, $$
\{\partial E - (\v_A)\cdot\mathbf{n}\mathbf{n} (\Delta t-s)\} - \{\partial E - (\v_B)\cdot\mathbf{n}\mathbf{n} (\Delta t-s)\}
$$ 
distributes the particles that should flow out but undo because of the jump. The total change is $$
- (\v_B - \v_A)\cdot\mathbf{n}(\Delta t-s)
$$

\item Finally, both $ \v_A $ and $ \v_B $ are pointing inward from the boundary $ \partial E $. When $ \v_B $ is longer than $ \v_A $, $$
\{\partial E - (\v_B)\cdot\mathbf{n}\mathbf{n} (\Delta t-s)\} - \{\partial E - (\v_A)\cdot\mathbf{n}\mathbf{n} (\Delta t-s)\}
$$ 
distributes the particles that flow in $ E $ but should have not flown in without the alternation of $ y(\cdot) $; when $ \v_B $ is no longer than $ \v_A $, $$
\{\partial E - (\v_A)\cdot\mathbf{n}\mathbf{n} (\Delta t-s)\} - \{\partial E - (\v_B)\cdot\mathbf{n}\mathbf{n} (\Delta t-s)\}
$$ 
distributes the particles that fail to flow into $ E $ because of the jump. 

Notice both Case 3 and Case 4 still reveal the result of spatial measure of change to $ E $ $$
- (\v_B - \v_A)\cdot\mathbf{n} (\Delta t-s)
$$
\end{enumerate}
Here we end proving Proposition \ref{Prop20230815}.\qedsymbol
\end{prof}

All the four cases reveal a result of change of quantity in $ E $: $$
\sum\limits_{\d s} \lambda \d s\int_{\partial E} (\Delta t - s)(-\v_B + \v_A)\cdot\mathbf{n}\rho(x,t)\d x
$$By the Gauss-Green formula this equals to$$ 
\sum\limits_{\d s} (\Delta t - s)\int_E \triangledown_x\cdot\left((\v_B(x,s) - \v_A(x,s))\rho(x,t)\right) \d x\lambda\d s
$$ 
Let the maximum partitioning length $ \|\d s\| $ go to zero, this writes to the integral $$
\int_E \int_0^{\Delta t} (\Delta t - s)\triangledown_x\cdot\left[(\v_B(x,s) - \v_A(x,s))\rho(x,t)\right]\lambda\d s\d x
$$ 
which equals to $$
\frac{\lambda}{2}\Delta t^2 \int_E V(x)\d x = o(\Delta t)
$$
where $$
V(x) = \max_s\big(\triangledown_x\cdot\left[(\v_B(x,s) - \v_A(x,s))\rho(x,t)\right]\big)
$$
Now we have given the strict proof that Case Two brings $ o(\Delta t) $, and thus the whole proof of the theorem is complete.\qedsymbol
\end{prof}

\section{Breadth-Oriented Reasoning}
It is natural to consider multiple depth-oriented reasoning to happen simultaneously. If it is allowed that they share their cognitive states, the paradigm is suggested as Table \ref{Breadth_O_thinking20230723}. The readers may compare with Figure \ref{Fig20230817_1}.
\begin{table}[htb]
\centering
\caption{Multiple depth-oriented thinking that share their cognitive states}
\label{Breadth_O_thinking20230723}
\begin{tabular}{cll}
\hline
time & reasoning & cognitive states\\
\hline
$t_1$:& $x_{11}(t_1),x_{12}(t_1)$,$\cdots$,$x_{1k(t_1)}(t_1)$ &$y_{11}(t_1)$,$\cdots$,$y_{1j(t_1)}(t_1)$\\
$t_2$:& $x_{21}(t_2)$,$x_{22}(t_2),$ $\cdots$,$x_{2k(t_2)}(t_2)$ &$y_{21}(t_2)$,$\cdots$,$y_{2j(t_2)}(t_2)$\\
$\vdots$ &$\cdots $                                       &$\cdots   $                          \\
$t_N$:& $x_{N1}(t_N)$,$x_{N2}(t_N)$,$\cdots$,$x_{Nk(t_N)}(t_N)$ &$y_{N1}(t_N)$,$\cdots$,$y_{Nj(t_N)}(t_N)$\\
\hline
\end{tabular}
\end{table}

In this section, we provide a basic probabilistic glimpse of the system illustrated in Figure. We add zero items at the end to ensure that the length of reasoning sequences and cognitive sequences remains the same:
\begin{align*}
&x(t_1;\omega) = \left( x_{11}(t_1),x_{12}(t_1)\cdots,x_{1k(t_1)}(t_1),0,\cdots,0\right) \\
&y(t_1;\omega)= \left(y_{11}(t_1),\cdots,y_{1j(t_1)}(t_1),0,\cdots,0\right) \\
&x(t_2;\omega) = \left( x_{21}(t_2),x_{22}(t_2),\cdots,x_{2k(t_2)}(t_2),0,\cdots,0\right) \\
&y(t_2;\omega)= \left(y_{21}(t_2),\cdots,y_{2j(t_2)}(t_2),0,\cdots,0\right) \\
&\vdots                                                        \\
&x(t_N;\omega) = \left( x_{N1}(t_N),x_{N2}(t_N),\cdots,x_{Nk(t_N)}(t_N),0,\cdots,0\right) \\
&y(t_N;\omega)= \left(y_{N1}(t_N),\cdots,y_{Nj(t_N)}(t_N),0,\cdots,0\right) 
\end{align*}
We consider continuous-time system $$
\{x(t;\omega)\},\{y(t;\omega)\}
$$
for mathematical simplicity in this section. For real-valued functions $ \psi $ and $ \phi $,$$
\frac{\d }{\d t}\E\Big[\psi(x(t;\omega)), \phi(y(t;\omega))\Big] = \E\Big[\Big( \psi(t),\widehat{\mathscr{A}}_{\psi(t)}(\phi(t))\Big) + \Big(\phi(t),\widehat{\mathscr{A}}_{\phi(t)}(\psi(t))\Big)\Big]
$$
where\begin{align*}
&\lim_{\Delta t\downarrow 0}\frac{1}{\Delta t}\E\Big[\psi(x(t+\Delta t))-\psi(x(t))\big\vert \psi(x(t)) = u,\phi(y(t) = w \Big] = \widehat{\mathscr{A}}_w(u) \\
&\lim_{\Delta t\downarrow 0}\frac{1}{\Delta t}\E\Big[\phi(y(t+\Delta t))-\phi(y(t))\vert \psi(x(t)) = u,\phi(y(t)) = w\Big] = \widehat{\mathscr{A}}_u(w)
\end{align*}
This is because\begin{align*}
&\E\Big[\Big(\psi(x(t+\Delta t)),\phi(y(t+\Delta t))\Big) - \Big(\psi(x(t)), \phi(y(t))\Big) \Big\vert \Big(\psi(x(t)), \phi(y(t))\Big) = (u,w) \Big] \\
=&\E\Big[\Big(\psi(x(t+\Delta t)),\phi(y(t+\Delta t))\Big) - \Big(\psi(x(t+\Delta t)), \phi(y(t))\Big) \Big\vert \Big(\psi(x(t)), \phi(y(t))\Big) = (u,w) \Big] \\
+&\E\Big[\Big(\psi(x(t+\Delta t)),\phi(y(t))\Big) - \Big(\psi(x(t)), \phi(y(t))\Big) \Big\vert \Big(\psi(x(t)), \phi(y(t))\Big) = (u,w) \Big] \\
=&\Big(u+O(\Delta t),\E\Big[\phi(y(t+\Delta t))-\phi(y(t))\vert \psi(x(t)) = u,\phi(y(t)) = w\Big]\Big)\\
+&\Big(\E\Big[\psi(x(t+\Delta t))-\psi(x(t))\vert \psi(x(t)) = u,\phi(y(t)) = w\Big],w\Big)\\
=& (u,\widehat{\mathscr{A}}_u(w))\Delta t + (\widehat{\mathscr{A}}_w(u),w)\Delta t + o(\Delta t)
\end{align*}
The probabilistic model is referred to as Markov processes with Markovian switching. When the system $ \big(\psi, \phi \big) $ contains Brownian process $ \B(t) $ with an order of $ \Delta t^{\frac{1}{2}} $, it writes:
$$
\d \big(\psi, \phi \big) = \left[\widehat{\mathscr{A}}_\psi(\phi) + \widehat{\mathscr{A}}_\phi(\psi) + \frac{1}{2}\widehat{D}\widehat{D}\right]\big(\psi, \phi \big)\d t 
+ \widehat{D}\big(\psi, \phi \big)\d \B(t)
$$

We follow the language used in quantum mechanics when considering $ \psi $ as an element of the complex Hilbert space. In the complex Hilbert space, given a standard orthogonal basis, $ \ket{u} $ represents the column vector projection of $ u $ onto this basis. The conjugate transpose of $ \ket{u} $ is depicted as $ \bra{u} $. Meanwhile, linear operators $ \widehat{\mathscr{A}}_\phi $,$ \widehat{D} $ are treated as matrices $ {\mathscr{A}}_\phi $, $ D $ relative to this standard orthogonal basis. The last formula is written as $$
\d \big[\ket{\psi}, \ket{\phi} \big] = \left(\left[I,{\mathscr{A}}_\psi\right] + \left[{\mathscr{A}}_\phi,I\right] + \frac{1}{2}{D}{D}\right)\big[\ket{\psi}, \ket{\phi} \big]\d t + {D}\big[\ket{\psi}, \ket{\phi} \big]\d \B(t)
$$
Here the square brackets are used as parentheses, and not as commutation in physics.

In the following text, our focus is solely on the reasoning states $ \ket{\psi} $ while disregarding the cognitive states. $$
\d \psi(t,\B(t))= \left[\widehat{\mathscr{A}}_{\phi(t)}  + \frac{1}{2}\widehat{D}\widehat{D}\right]\psi(t,\B(t))\d t + \widehat{D} \psi(t,\B(t))\d\B(t)
$$ 
is written as$$
\d \ket{\psi(t)} = \left[{\mathscr{A}}_{\phi(t)}  + \frac{1}{2}{D}{D}\right]\ket{\psi(t)} \d t + D\ket{\psi(t)}\d \B(t)
$$
Consider $ k $ states $ \ket{\phi_1},\cdots, \ket{\phi_k} $, the density matrix is defined as
$$
\rho = \sum_{j=1}^k \ket{\psi_j}\bra{\psi_j}
$$ 
We have
\begin{equation}\label{20230720MyLindblad_prev}
\d\rho = \Big( {\mathscr{A}}_{\phi(t)}\rho + \rho {\mathscr{A}}_{\phi(t)}^\dagger -\frac{1}{2}D^\dagger D\rho-\frac{1}{2}\rho D^\dagger D + D\rho D^\dagger \Big)\d t + \Big(D\rho + \rho D^\dagger\Big)\d \B(t)
\end{equation}
where $ \dagger $ is to take conjugate transpose of matrices, $$
D^\dagger = -D
$$
Taking expectation of both sides in \eqref{20230720MyLindblad_prev}, $ \rho_\E = \E\rho $, we have \begin{equation}\label{20230720MyLindblad}
\d\rho_\E = \Big( {\mathscr{A}}_{\phi(t)}\rho_\E + \rho_\E{\mathscr{A}}_{\phi(t)}^\dagger -\frac{1}{2}D^\dagger D\rho_\E -\frac{1}{2}\rho_\E D^\dagger D + D\rho_\E D^\dagger \Big)\d t 
\end{equation}
Please note that the format of our result is resembled to the Lindblad master equation, and we cite\cite{caruso2016fast} and \cite{mccauley2020accurate} for reference.

\bibliographystyle{unsrt}
\bibliography{templateArxiv.bib}

\end{document}